\documentclass[11pt,a4paper]{article}
\usepackage[hyperref]{acl2020}
\usepackage{times}
\usepackage{latexsym}
\usepackage{graphicx}

\usepackage[linesnumbered,ruled,vlined,noend]{algorithm2e}
\usepackage{setspace}

\usepackage{microtype}

\aclfinalcopy % Uncomment this line for the final submission

\title{JIT-Masker: Efficient Online Distillation for Background Matting}

\author{Jo Chuang, Qian Dong \\
  Stanford University \\
  \texttt{\{jochuang, qiandong\}@stanford.edu}}

\date{}

\begin{document}
\maketitle

\begin{abstract}
    We design a real-time portrait matting pipeline for everyday use, particularly for "virtual backgrounds" in video conferences. Existing segmentation and matting methods prioritize accuracy and quality over throughput and efficiency, and our pipeline enables trading off a controllable amount of accuracy for better throughput by leveraging online distillation on the input video stream. We construct our own dataset of simulated video calls in various scenarios, and show that our approach delivers a 5x speedup over a saliency detection based pipeline in a non-GPU accelerated setting while delivering higher quality results. We demonstrate that an online distillation approach can feasibly work as part of a general, consumer level product as a "virtual background" tool. Our public implementation is at \href{https://github.com/josephch405/jit-masker}{https://github.com/josephch405/jit-masker}.
\end{abstract}

\section{Background}

With the recent surge in popularity of online video conferencing tools, "virtual backgrounds" have become an interesting cultural phenomenon. For every input video frame, a real-time system classifies pixels into either the foreground (in most cases, representing the user) or background. Users select an image or video that replaces the background pixels, which is composited with the foreground layer to generate an artificial video stream. Historically, using a green screening was the best method for achieving high quality matting - however, the setup involved in doing so is impractical for the average consumer. Recent commercial offerings are capable of masking out backgrounds in regular video conference feeds without the use of a green screen.

The three main goals and design principles we define to be important for a good "virtual background" system are as follows:

1. \textbf{Speed}: The system needs to process and infer each frame fast enough to keep a live video smooth. If necessary, it is acceptable to compensate video quality or accuracy for throughput. As we are only every working on a singular stream, we do not have to consider the effects of batching on latency, and for the most part throughput is inversely correlated to latency. This speed should be transferrable to settings where there is limited compute, ie. on a laptop or mobile device.

2. \textbf{Accuracy}: The model is able to reasonably separate the salient object from the background. In the majority of cases this will be a person, but this is not a hard requirement in some situations. As noted from before, we can accept "good enough" in exchange for consistent throughput. Additionally, we should be able to control this tradeoff between accuracy and speed.

3. \textbf{Memory}: Our main pipeline should work without an excessive amount of memory usage. It should be able to work on a general laptop, and if possible, on mobile devices.

Our main intuition behind constructing the JIT-Masker pipeline is inspired by the JITNet approach from \cite{mullapudi2019online}, which uses online model distillation to speed up inference while maintaining relatively high accuracy. The main questions we set out to investigate shifted over the course of the project, but in general we stayed focused on the following:

1. Is it feasible to build a neural "virtual background" pipeline, performant on laptops and mobile devices? Currently available consumer solutions (ie. from Zoom) for virtual backgrounds are quite resource efficient - however, they face certain limitations such as inferior fine-grained quality and the inability to run on devices the applications deems "incompatible" (ie. laptops below a certain level of compute capability, any Linux machine, mobile phones). In contrast, traditional neural network approaches tend to exploit the massive parallelism of GPU devices or specialized compute units, which are not generally available. 

2. Are there improvements we can make in recognizing previously encountered scenes? As our primary approach involves fine-tuning on a live stream, we should strive to avoid spending compute on repeated work if possible. This was a particular point that the original JITNet paper did not attempt to address directly, and we would like to explore this issue in our pipeline.

\section{Related work}
\subsection{Model Distillation}
The practice of training a smaller student network to match the predictions of a larger teacher network has been well explored in the context of deep learning. For a wide variety of tasks, a smaller model distilled from a larger network often outperforms the model trained on the same training data \citep{Hinton2015DistillingTK}. While the student model cannot achieve the same performance as the teacher, the higher performance gained from distillation enables high performing models in low resource settings \citep{Howard2017MobileNetsEC, sanh2019distilbert}.

\subsection{Online model distillation}
\citep{mullapudi2019online} propose JITNet, a video segmentation framework that exploits the temporal coherence between frames to reduce computation cost and leverages a high-quality teacher model to perform online distillation. This distillation approach is employed to take advantage of the fact that most video streams observe a very small subset within the general distribution of real-world images (eg. a fixed corner of a traffic crossing, one particular room), and that we can achieve high-enough quality predictions with massive reductions in compute cost.
\subsection{Salient Object Detection}
Saliency is the task of segmenting the most visually attractive objects in a scene. Most recent work has focused on refining the visual quality of the predictions. \cite{qin2020u2} proposes U$^2$Net, a U-Net architecture utilizing efficient pooling and residual layers. They also introduce a U$^2$Net$^\dagger$ variant that is significantly smaller (model weights are 4.7 MB vs 176.3 MB for U$^2$Net) but still on par with state-of-the-art performance.

% \subsection{Alpha Matting}
% Alpha matting 

% also approaches moving away from

% \subsection{Portrait Segmentation}

\section{Datasets}

In order to evaluate different approaches, we utilize datasets for both traditional segmentation as well as video conferencing contexts. 

\subsection{DAVIS}
The DAVIS 2016 dataset from \cite{perazzi2016benchmark} is a video saliency dataset spanning four evenly distributed classes including humans, animals, vehicles and objects. It contains 50 videos and 3455 frames in total with pixelwise labels for a single foreground object.

\subsection{Supervisely Person Dataset}
The Supervisely Person Dataset in \cite{supervisely} is a person image dataset with high quality annotations. The labels are genereated by Faster-RCNN and UnetV2 neural networks with manual validation and correction. The dataset consists of 5711 person images in total.

\subsection{VideoCall Dataset} We created a new dataset dedicated to the evaluation of our pipeline. There are 17 recorded zoom videos split across easy, medium and hard scenarios with 7, 6 and 4 videos in each respective category. All videos are formatted to 480p and cropped to 1 minute in length in order to capture sufficient variation properly representing a real call.

We define the difficulty to be directly tied to the number of scene changes in the video. A scene change is defined as a dramatic change in lighting, persons in the video (appearing or disappearing), or the background scene. Easy videos have no scene changes (ie. a person talking and staying within a still camera frame). Medium videos have one scene change, and hard videos include two or more scene changes.

\section{Proposed Method: JIT-Masker}

\begin{figure*}[ht]
    \centering
    \includegraphics[width=\textwidth]{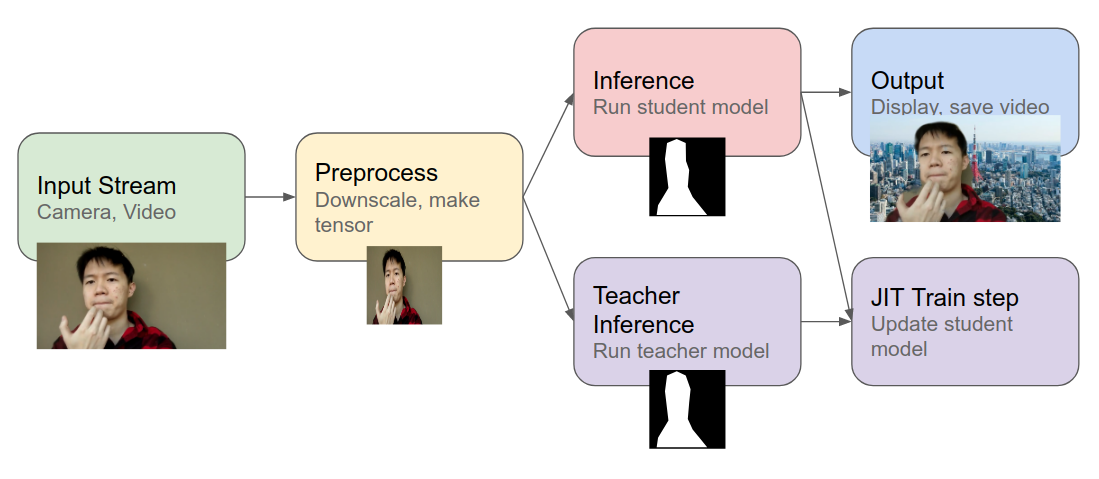}
    \caption{General structure of JIT-Masker. Each block seen here roughly corresponds to a separate thread in our pipeline}
    \label{fig:pipeline}
\end{figure*}

We propose JIT-Masker, an end-to-end pipeline for performing virtual background generation. The formal definition of the virtual background task can be defined as follows:

The inputs will be a stream of RGB video frames, $X_t \in [w, h, 3]$. For each frame, our system must produce $\alpha_t \in [w, h]$ that segments the subject of the video. A combined image $Y_t = (1 - \alpha_t) X_t + \alpha_t * B$ where $B \in [w, h]$ is the static background. The end-to-end pipeline is responsible for all the intermediate steps, as described in figure~\ref{fig:pipeline}. Working with the end to end pipeline ensures that we are working on the most critical part of the pipeline at all times and not simply optimizing subgoal metrics such as model inference time, independent of the rest of JIT-Masker.

\subsection{IoU-Acc Metric}
\label{sec:iou}
An interesting caveat for our system is that we need to handle situations where there are no positive examples in the input, ie. a background image with no person. The naive definition of the Intersection over Union metric is poorly defined in these cases, as the intersection and therefore IoU metric would always be zero. This penalizes models that correctly predict less area in empty frames as the IoU metric does not capture performance in these frames at all.

We propose a more lenient definition of IoU that we will call IoU-Acc: if the ground truth area consists of less than 5\% of the input area, we replace the IoU metric with accuracy of the prediction over the entire frame. We will explain the implications of this revised IoU in section~\ref{sec:distsched}.

\subsection{Student network: JITNet model}

\begin{figure}[tb] \centering
\includegraphics[width=\linewidth]{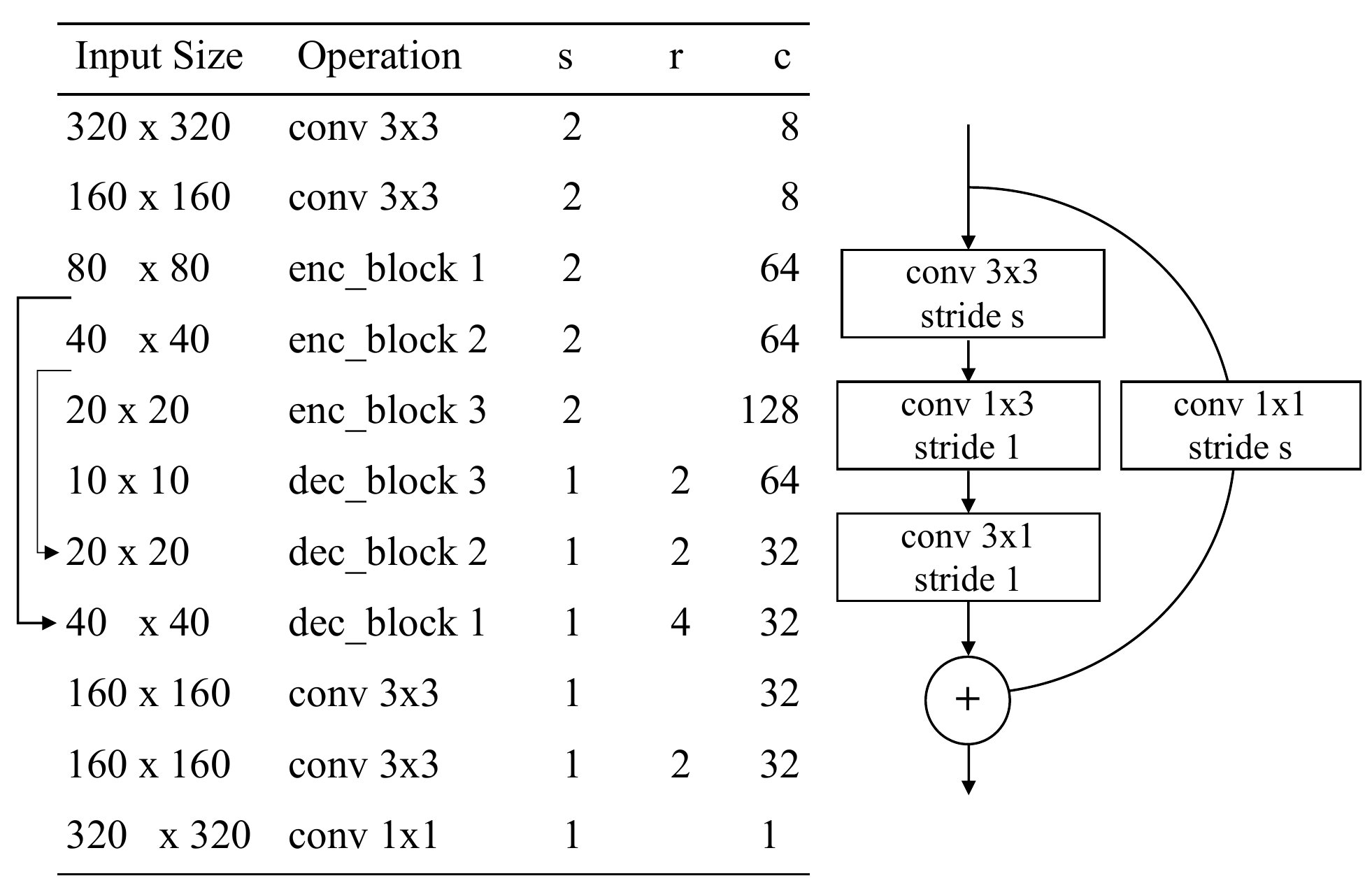}
\vspace{-2em}
\caption{Left: Our JITNet architecture. Right: encoder/decoder block details. s = stride, r = resize, c = output channels. Our JITNet handles smaller images that the original version. The network shown handles 320p inputs; only the spatial dimensions are changed for the 240p version of JITNet.}
\label{fig:architecture}
\vspace{-1em}
\end{figure}

While we can pick any arbritrary model as the student model, we opted for using the original JITNet model from \citep{mullapudi2019online}. We modify the network to output one channel as output that predicts the alpha mattes $\alpha^t$. Also, we fixed our networks to operate on downsized inputs for efficient inference - we found that downsampling $X_t$ before passing it into any CNN network was the most efficient way to save on inference time, while only moderating sacrificing quality. The specific layout of our JITNet is shown in figure~\ref{fig:architecture}.

We pretrain our JITNet model on the \citep{supervisely} dataset of human segmentations with a straightforward regime of BCE Loss and $Adam(lr=0.001, betas=(0.9, 0.999, eps=1e-08))$. This is to ensure that the model still outputs reasonable masks at the start of the stream without additional training from the teacher. Additionally, this is a form of regularizing the model by initializing on a better prior. Without pretraining, the student tends to overfit to the initial stream and struggles to adapt to later scene changes.

\subsection{Teacher Network: MRCNN}

Again, we can pick any arbitrary teacher model depending on the specific goals of the pipeline. Given the relevance of Salience Detection to our goal (ie. find the important object in the frame), we attempted to use U$^2$Net from \citep{qin2020u2} as a teacher, in addition to the more conventional choice of Detectron/MRCNN from \citep{wu2019detectron2}.

We validate the quality of each approach by directly comparing the Intersection over Union (IoU) score of predictions on the DAVIS \citep{perazzi2016benchmark} and Supervisely dataset \citep{supervisely}, as shown in Table~\ref{tab:teacher}. While DAVIS does not exclusively contain video clips of humans and the Supervisely dataset is an image dataset, we believe a joint evaluation on both would be a fair representation of performance on the distribution of video conferencing videos.

\begin{table}[ht]
    \small
    \centering
    \begin{tabular}{|c|c|c|c|c|}
        \hline
        & MRCNN50 & U$^2$Net & U$^2$Net$^\dagger$ \\
        \hline \hline
        DAVIS & .698 & \textbf{.742} & .732 \\
        \hline
        Supervisely & \textbf{.836} & .721 & .680 \\
        \hline
    \end{tabular}
    \caption{IoU results of various teacher models.}
    \label{tab:teacher}
\end{table}

The results indicate that while U$^2$Net was slightly better at saliency detection in videos, MRCNN50 vastly outperformed on person segmentation. In light of this information, we select MRCNN50 as our teacher in most JIT-Mask experiments unless indicated otherwise. This proved to be the correct choice as U$^2$Net was a lot less temporally consistent when evaluated on videos.

\subsection{Distillation schedule}
\label{sec:distsched}

We adapt most of the original JITNet distillation algorithm, with a few major caveats for the sake of performance and quality. Our modified variant of the original JITNet algorithm is presented in Algorithm~\ref{alg:online_distillation}, with $u$ indicating training budgets, $\delta$ indicating teacher inference intervals, $a_{thresh}$ indicating a desired score threshold, and $\theta$ representing the student network parameters.

\begin{algorithm}
\setstretch{0.90}
\SetAlgoLined
\DontPrintSemicolon 
\SetKwInput{KwInput}{Input}
\SetKw{True}{true}
\SetKw{False}{false}
\SetKw{And}{and}
\KwInput{$S_{0\dots n}$, ${u}_{max}$, ${\delta}_{min}$, ${\delta}_{max}$, ${a}_{thresh}$, $\theta$}
$\delta$ $\leftarrow$ ${\delta}_{min}$ \;
$a_{curr} = 0$ \;
\For{$t$ $\leftarrow$ 0 \KwTo n } {
    $update \leftarrow a_{curr} < a_{thresh}$ \;
    \If{ $t \equiv 0 \pmod{\delta}$} {
        $L_t$ $\leftarrow$ Teacher($S_t$) \;
        $u$ $\leftarrow$ $u_{max}$\;
        $a_{curr}$ $\leftarrow$ IoU-Acc($L_t$, $P_t$) \;
    }
    \ElseIf {update \And $u > 0$} {
        $\theta$ $\leftarrow$ UpdateStudent($\theta$, $P_t$, $L_t$) \;
        $P_t$ $\leftarrow$ Student($\theta_{t}$, $S_t$) \;
        $a_{curr}$ $\leftarrow$ IoU-Acc($L_t$, $P_t$) \;
        $u$ $\leftarrow$  $u - 1$ \;
    } 
    \ElseIf {update \And $u == 0$} {
        $\delta$ $\leftarrow$ max($\delta_{min}$, $\delta / 2$)
    }
    \ElseIf {not update \And $u > 0$} {
        $\delta$ $\leftarrow$ min($\delta_{max}$, $2 \delta$) \;
        $u$ $\leftarrow$ $0$
    }
}
\caption{JIT-Mask Online distillation}
\label{alg:online_distillation}
\end{algorithm}

First, given the structure of JIT-Masker, we break out the distillation process into its own thread independent of the main pipeline, ie. the student inference workflow. This teacher thread asynchronously updates the student weights. By doing so, we avoid having the main thread lock up while we run teacher inference and distillation.

Second, we only ever perform one operation per input image receive by the teacher. These operations are one of predicting teacher outputs, running a single learning step, or setting $\delta$ to an appropriate value. This mitigates a particular issue with the original distillation algorithm where up to $u_{max}$ student learning updates can happen before the next video frame is processed, which leads to a noticeable "freezing" phenomenon whenever we trigger learning. By spreading out the updates across each video frame, we are running the same update steps but issuing them across time.

Finally, we use our new IoU-Acc metric from Section~\ref{sec:iou}. This prevents our network from aggressively learning to fit empty images with no persons in the frame. Correspondingly, the inference pipeline always outputs the pure background frame whenever the student network predicts a mask that has an area less than the defined threshold (5\%). Without this specialized metric and inference heuristic, the student network often overfits to predict empty outputs, then struggles to recognize the user once she or he re-enters the frame.

For our experiments, we set $u_{max}=8$, $\delta_{min}=8$, $\delta_{max}=64$ and $a_{thresh} = 0.9$. We train the student model with Stochastic Gradient Descent and a learning rate of 0.2. Also, unlike the original JITNet, we do not downweight the loss on background areas of the image since for most inputs our class distribution is fairly well balanced between the foreground and background.

\section{Results}

We evaluate all results on our VideoCall dataset with the 240p variant of the pipeline. The two quantitative metrics we measure are \textbf{Quality}, as represented by IoU-Acc with respect to MRCNN50 "ground-truth" predictions, as well as \textbf{Speed}, represented by the average time between "paints" on the final output stage of the pipeline.

We run our pipelines both on GPU and CPU. When we run on the CPU, only student model inference and distillation are run on the CPU and not teacher inference. This is to emulate a potential real-world setup where low-power consumer devices can send teacher inference requests to a remote endpoint, instead of having to run the teacher themselves.

Our baseline comparison is a naive approach of taking a pretrained U$^2$Net$\dagger$ and directly passing all frames through the network to predict the masks. These results are shown in Table~\ref{tab:res}. We select U$^2$Net$\dagger$ as a reference given that it also targets limited compute settings.

All experiments were run on a machine equipped with a GTX 1080 GPU and an Intel Quad-Core i5 7600K CPU @ 3.80 GHz.

\begin{table}[t]
    \small
    \centering
    \begin{tabular}{|c|c|c|c|c|}
        \hline
        Model & IoU-Acc & GPU ms & CPU ms \\
        \hline \hline
        JIT-Masker & \textbf{.8950} & \textbf{40}  & \textbf{83} \\
        \hline
        U$^2$Net$\dagger$ & .8326 & 44 & 447 \\
        \hline
        MRCNN50 & - & 91 & - \\
        \hline
    \end{tabular}
    \caption{Performance of our approach against non-distillation approaches}
    \label{tab:res}
\end{table}

Results indicate that our model is relatively close to predicting the same outputs as the teacher, at least a lot more than a pretrained saliency detector. Additionally, we are doing so at a significantly lower cost, up to 5x faster than the lightweight U$^2$Net$\dagger$ on CPU. With 83 ms per frame, we can comfortably process 10 frames per second (FPS) even on CPU, before even considering lower level optimizations that we did not perform on our Python-based pipeline. As modern video conferencing platforms typically run at a framerate of 10 FPS or less, this strongly suggests that we can work towards a decent production-level solution using online distillation, provided that we optimize the pipeline further.

\subsection{Difficulty vs. Efficiency}

\begin{figure}[tb] \centering
\includegraphics[width=\linewidth]{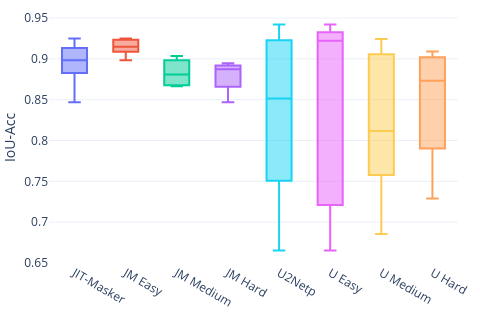}
\vspace{-2em}
\caption{IoU-Acc score vs. Model and Difficulty Split}
\label{fig:perf}
\vspace{-1em}
\end{figure}

\begin{figure}[tb] \centering
\includegraphics[width=\linewidth]{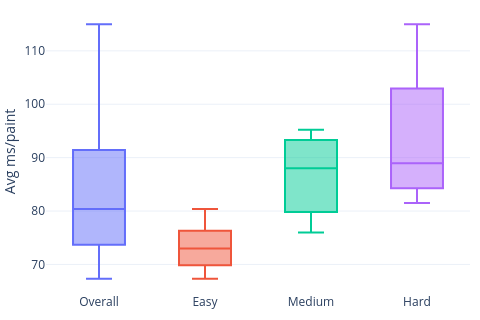}
\vspace{-2em}
\caption{Avg ms / frame vs. Difficulty of Videos}
\label{fig:speed}
\vspace{-1em}
\end{figure}

We further broke down the performance of our models across the difficulty of the videos, as shown in Figures~\ref{fig:perf} and ~\ref{fig:speed}.

Overall, we see that JIT-Masker segments easier videos with higher accuracy and less time. This validates our intuition that with more scene changes (which define our levels of difficulty), the model needs to spend more time on adapting to the solution. However, the pipeline actually spends a non-trivial amount of time simply pre-processing and post-processing, as shown in Table~\ref{tab:breakdown}. This strongly suggests that attempts at reducing delay by recognizing previous scenes, which was our second line of inquiry, would most likely be overshadowed by simple improvements to other parts of the pipeline, including skipping resizing the inputs and masks. 

\begin{table}[ht]
    \small
    \centering
    \begin{tabular}{|c|c|c|c|c|}
        \hline
        Component & ms \\
        \hline \hline
        Camera & 7 \\
        \hline
        Preprocess & 35 \\
        \hline
        Student Inference & 9 \\
        \hline
        Output & 27 \\
        \hline
    \end{tabular}
    \caption{Breakdown of processing times for each component of the critical path of the pipeline. Measurements done on GPU JIT-Masker.}
    \label{tab:breakdown}
\end{table}

\section{Conclusion}
We proposed JIT-Masker, a virtual background pipeline based on online distillation of a student model towards a teacher model. By distributing work via threads and conducting distillation asynchronously alongside the main inference task, we demonstrate the feasibilty of developing an online neural approach to the "virtual background" task.

% ~\citep{Gusfield:97} \citet{Gusfield:97}
% \section*{Acknowledgments}

\bibliography{acl2020}
\bibliographystyle{acl_natbib}
\appendix

% \section{Appendices}
% \label{sec:appendix}

% \section{Supplemental Material}
% \label{sec:supplemental}

\end{document}